\documentclass[manuscript]{acmart}

\AtBeginDocument{%
  \providecommand\BibTeX{{%
    \normalfont B\kern-0.5em{\scshape i\kern-0.25em b}\kern-0.8em\TeX}}}

\copyrightyear{2021}
\acmYear{2021}
\setcopyright{acmcopyright}\acmConference[iiWAS2021]{The 23rd International
Conference on Information Integration and Web Intelligence}{November 29-December
1, 2021}{Linz, Austria}
\acmBooktitle{The 23rd International Conference on Information Integration and Web
Intelligence (iiWAS2021), November 29-December 1, 2021, Linz, Austria}
\acmPrice{15.00}
\acmDOI{10.1145/3487664.3487728}
\acmISBN{978-1-4503-9556-4/21/11}

\begin{document}

\title{A Generic Knowledge Based Medical Diagnosis Expert System}

\author{Xin Huang}
\email{xinh1@umbc.edu}
\affiliation{%
  \institution{University of Maryland, Baltimore County}
  \country{USA}
}

\author{Xuejiao Tang}
\email{xuejiao.tang@stud.uni-hannover.de}
\affiliation{%
  \institution{Leibniz University of Hannover}
  \country{Germany}
}

\author{Wenbin Zhang}
\email{wenbinzhang@umbc.edu}
\affiliation{%
  \institution{University of Maryland, Baltimore County}
  \country{USA}
}

\author{Shichao Pei}
\email{shichao.pei@kaust.edu.sa}
\affiliation{%
  \institution{King Abdullah University of Science and Technology}
  \country{Saudi Arabia}
}

\author{Ji Zhang}
\email{ji.zhang@usq.edu.au}
\affiliation{%
  \institution{University of Southern Queensland}
  \country{Australia}
}


\author{Mingli Zhang}
\email{mingli.zhang@mcgill.ca}
\affiliation{%
  \institution{Mcgill University}
  \country{Canada}
}

\author{Zhen Liu}
\email{liuzhen@gdufs.edu.cn}
\affiliation{%
  \institution{Guangdong University of Foreign Studies}
  \country{China}
}

\author{Ruijun Chen}
\email{n78083016@mail.ncku.edu.tw}
\affiliation{%
  \institution{National Cheng Kung University}
  \country{China}
}

\author{Yiyi Huang}
\email{yiyi063@email.arizona.edu}
\affiliation{%
  \institution{University of Arizona}
  \country{USA}
}

\renewcommand{\shortauthors}{X. Huang, X. Tang,  W. Zhang, S. Pei, J. Zhang et al.}

\begin{abstract}
In this paper, we design and implement a generic medical knowledge based system (MKBS) for identifying diseases from several symptoms. In this system, some important aspects like knowledge bases system, knowledge representation, inference engine have been addressed. The system asks users different questions and inference engines will use the certainty factor to prune out low possible solutions. The  proposed  disease  diagnosis  system  also uses  a  graphical  user interface (GUI) to facilitate users to interact with the expert system. Our expert system is generic and flexible, which can be integrated with any rule bases system in disease diagnosis.  
\end{abstract}



\maketitle

\section{Introduction}


AI systems are now widely used in economics, medicine, engineering and social medias, and are also built into many software applications and computer strategy games such as computer chess and computer video games. Expert system can process large amounts of known information and apply reasoning capabilities to provide conclusions. An expert system is a system that employs human knowledge captured in an automated system to solve problems that typically require human expertise. 


In this paper we propose the design and development of a medical knowledge based system (MKBS) for disease diagnosis from symptoms. It provides rich features for searching properties like symptoms, treatments, hierarchical clusters of particular diseases. The system supports a knowledge construction module and an inference engine module. The knowledge construction was built on a concept of rules, which was represented in a tree structure, and properties of a particular disease were stored as a semantic net. The inference engine uses the interactive backward chaining technique to infer a diagnostic result and. This medical expert system has been developed using JIProlog interpreter and Java - which provides an interactive user interface by asking questions and a visualization module that can be used by inference engine to produce visual decision making tree. The latter one can be used by explanation subsystem for visualizing queries. Furthermore, this system has a knowledge base editor module to dynamically add, edit, and drop rules and their premises. 

Our Diagnosis Expert System's knowledge base has 23 diseases from different categories, as shown in Table~\ref{tbl_disease}. The user is asked to answer with certainty factor, Yes or No if a certain symptom appears or not. Based on the user’s answers and the reasoning of the inference engine, the name of the disease is posted on the screen along with its certainty factor. 
\begin{table*}[]
\caption{Diseases diagnosed in our system.}\label{tbl_disease}
\begin{tabular}{|l|l|l|l|l|}
\hline
Cancer &  Anxiety Disorders & Balding and hair loss & Thyroid disorders & Heart attack\\
\hline
Diabetes &  Asthma & Erectile disorders & Migraine & Heart disease\\
\hline
Allergies &  Prostate conditions & Lupus & Skin disorders & Chest Pain\\
\hline
Cohn’s Disease &  Abdominal Pain & Thyroid disorders & Eye disorders & Respiratory problem\\
\hline
Common Cold &  Anxiety & Impotence & Headache & Abdominal Pain\\
\hline
Diarrhea &  Tonsil & Speech problem & &\\
\hline
\end{tabular}
\end{table*}

The contributions of the proposed medical diagnosis expert system are:
\begin{itemize}
    \item Build a medical expert software system for identifying diseases by several symptoms and describing methods of treatment to be carried out.
    \item Use MKBS system to diagnose a major disease by providing symptoms or get properties information about medical diseases.
    \item Provide a generic GUI  visualization tool for any rule based expert system build on core exshell. Users can use this visualization tool to understand how the system making decisions about a disease.
    \item Provide a knowledge based GUI editor to dynamically add, update or delete rules in current working memory.
\end{itemize}

\section{Related Work}
Expert Systems are computer programs that are derived from  Artificial Intelligence (AI). AI’s scientific goal is to understand intelligence and gain insights by building computer programs that exhibit intelligent behavior~\cite{ref_luger}. It contains concepts and methods of symbolic inference, or reasoning by a computer, and the mechanism of using the knowledge to make inferences  inside the machine.

The development of traditional expert systems was aided by the development of the symbolic processing languages Lisp and Prolog~\cite{ref_merritt}. To avoid reinventing the wheel, expert system shells were created with specialized features for building large expert systems. 

Recent advances in the field of artificial intelligence have led to the emergence of medical expert systems~\cite{ref_salem}~\cite{ref_riche}. Successful expert systems MYCIN was an early expert system designed to identify bacteria causing severe infections, such as bacteremia and meningitis, and to recommend antibiotics treatment~\cite{ref_Wiri}. Developing a reliable medical diagnosis system is a complex task. Recently several fuzzy logic based expert systems have been developed for  diagnosis of specific diseases for coronary artery disease~\cite{ref_muham} and chronic kidney disease~\cite{ref_jimmy}. 

In this paper, we aim to develop a generic knowledge based medical expert system. This medical expert system has the scope of representing some major diseases' symptoms and properties. One of the main focuses of this medical system is to develop graphical user interface to visualize decision making and provide a knowledge-based GUI editor. 

\begin{figure*}[ht]
\includegraphics[scale=0.6]{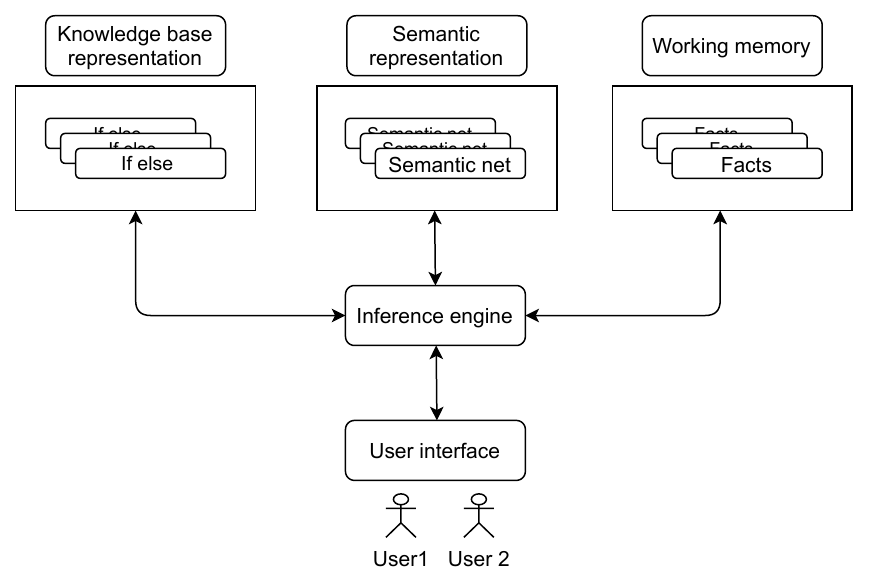}
\caption{High level design of knowledge based disease diagnosis expert system.} \label{fig_system}
\end{figure*}

\section{An Expert Framework for Disease Diagnosis}
An expert system is composed of two major parts, a knowledge base and an inference engine. The inference engine reasons about the knowledge base like a human to run an expert system. A dialog interface is used to communicate between users and the system, which has ability to conduct a conversation with users via "conversational" interface. 

Our medical expert system is built with typical  components of an expert system. To introduce a visualization module in exshell, this system uses Java, JIProlog – a java implementation of prolog~\cite{ref_jiprolog} and visualization tool Yfiles~\cite{ref_yfile}.
Figure\ref{fig_system} shows the high level design of our knowledge based disease diagnosis expert system. This system has all of the necessary components that an expert system must have. The major components are:
\begin{itemize}
    \item Knowledge base – A declarative representation of the expertise, can either be the IF-THEN rules based or knowledge base representation such as semantic net and frame based representation. 
    \item Inference engine – It is the core of the system, which derives recommendations from the knowledge base and problem-specific data in working memory.
    \item Working memory – It is used to store user inputs and for further matching queries, which help to not ask duplicate questions. 
    \item User interface – It is the interface that controls the interaction between the user and the core expert system. This component has been built using JIProlog and Java.
\end{itemize}


\subsection{Knowledge structure}
The knowledge construction was built on a concept of rules that were performed in a  hierarchical tree structure. For this part, rule based knowledge with a hierarchical tree structure is used to represent the knowledge about different diseases and its symptom.

This system has its own internal rule format with header information and all premises are connected by AND or OR conditions. This format has a section for representing a certainty factor of rules. This certainty factor will be used by the inference engine to make decisions about a possible solution. The rule can be nested, which means a rule can have nested sub-goal rules and different facts. 



\subsection{Inference Engine}

An inference engine is used to derive answers from the knowledge base. A finite state machine in the inference engine can be described with a cycle of three action states as match rules, select rules, and execute rules. Rules are designed to be a tree like hierarchical structure to represent the knowledge about different diseases and their symptoms. With the first state of match rules, the inference engine searches all of the rules that are satisfied by the current contents of the data target. In the second state, the inference engine applies some selection strategy to determine which rules will actually be executed. Finally, the inference engine executes or fires the selected rules, with the instantiating data items as parameters.

Using this format of rules, the inference engine of this system  can reason knowledge about diseases by several symptoms. The main desired behaviors of inference engine are: 
\begin{itemize}
    \item combine certainty factors as indicated previously. 
    \item maintain working memory that is updated as new evidence is acquired.
    \item locate all information specific to an attribute when it is asked for, and store that information in working memory.
\end{itemize}

\textbf{Backward chaining  approach}

The system works backward from the ultimate goal until all the sub-goals in working memory are known to be true, indicating that the hypothesis has been verified. This process is known as “backward chaining”. 

\textbf{Steps in backward chaining:} 
System has hypothetical solution(s) (e.g. “The patient has type I diagnosis(X)”), and tries to prove it.

\begin{itemize}
    \item Find rules that match the goal. 
    \item If antecedents match the facts, stop.
    \item If not, make the antecedents the new sub-goals, and repeat.
\end{itemize}

\textbf{Certainty Factors:}
The most common scheme for modeling uncertainty is to assign a certainty factor to each piece of information in the system. The inference engine automatically updates and maintains the certainty factors as the inference proceeds. The certainty factors ($CF$) are integers from 0 (for definitely false) to 1 (for definitely true). The rule format also allows for the addition of certainty factors. 



When the premise of a rule is uncertain due to uncertain facts, and the conclusion is uncertain due to the specification in the rule, the following formula is used to compute the adjusted certainty factor of the conclusion: 
\begin{equation}
CF = RuleCF * PremiseCF
\end{equation}

\subsection{Working memory}
One can treat working memory as the brain processes used for temporary storage and manipulation of information. It operates over a short time window which is usually a few seconds, and it facilitates users to focus attentions, resist distractions, and guide the decision-making. 

Working memory simply contains the known facts about attribute-value pairs. Known facts, represented as $Known(Goal, CF)$, are stored in the Prolog database for further use.
If a current fact and its certainty fact have been already stored in working memory, the system would not ask users about it further and it will  calculate certainty fact using the stored data.

\begin{figure*}[ht]

\includegraphics[width=\textwidth]{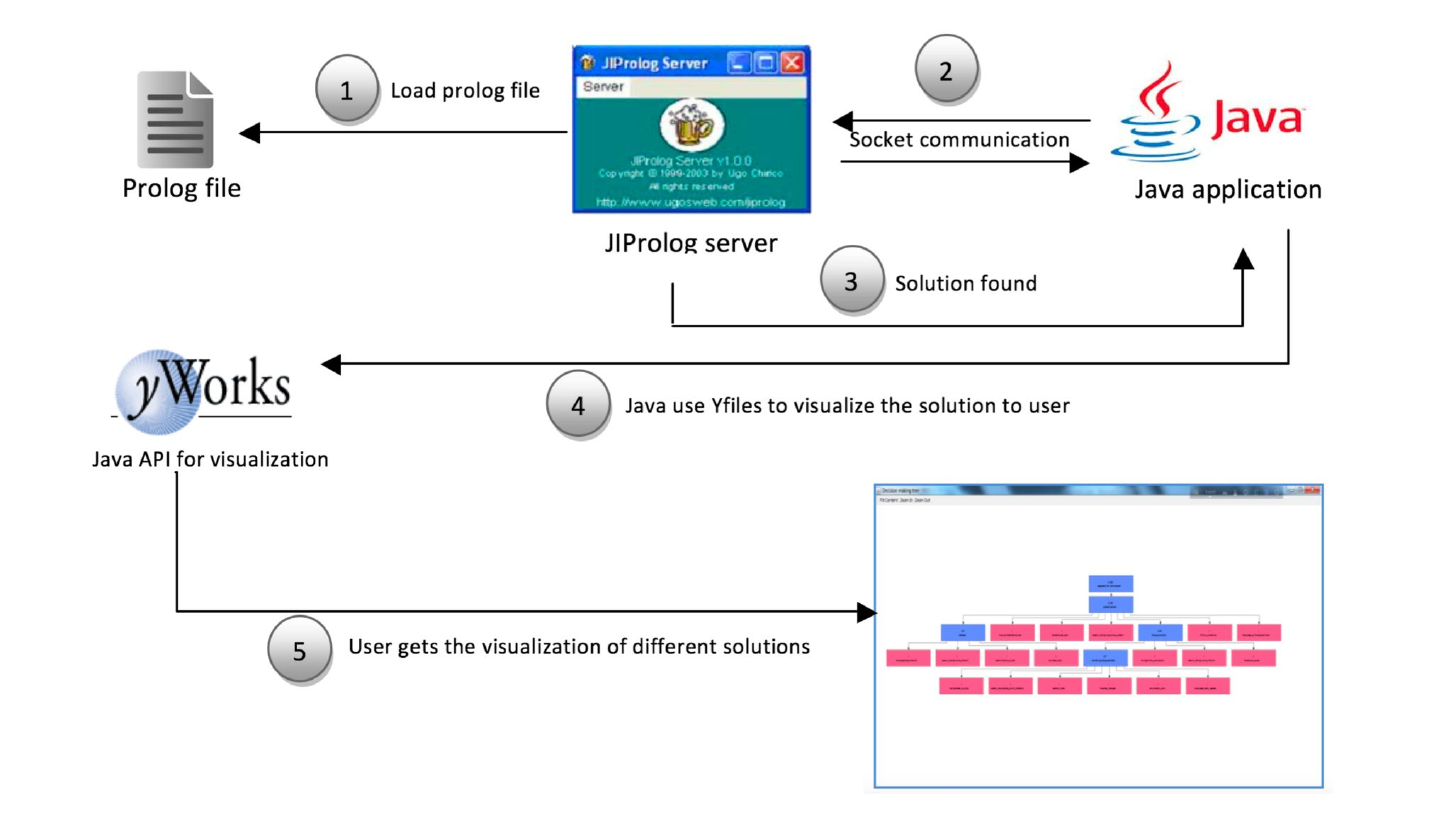}
\caption{End to end implementation of our disease diagnosis system.} \label{fig_demo}
\end{figure*}

\subsection{User interface}
One of the important contribution in our expert system is introducing a user interface to facilitate the visualization of decision making process. Traditionally, most of the prolog program’s output are texts, which are difficult for users to understand the solution provided by expert system. To provide a user friendly interface for the expert system, we develop a graphical user interface that can be used in any rule based system that is built on top of the core exshell system. This user interface is a generic interface which can be easily plugged into any rule based system built on top of core exshell.





\section{Framework Implementation}
The implementation of our disease diagnosis systems is shown in Fig.~\ref{fig_demo}. One of the main goals of this system is to introduce a  user friendly user interface for such a rule based expert system. With the combination of Java and prolog, we develop a complete end to end GUI based expert system for disease diagnosis.  Three main technologies used are:
\begin{itemize}
    \item Java: Java is a widely used high level programming language.
    \item JIProlog:  Java Internet Prolog is a cross-platform pure Java 100\% Prolog interpreter that integrates Prolog and Java languages in a very fascinating way. JIProlog allows calling Prolog predicates from Java without dealing with native code (JNI). JIProlog can consume any prolog file that has ISO standard prolog syntax.
    \item Yfiles:  YFiles for Java is an extensive Java class library that provides algorithms and components enabling the analysis, visualization, and the automatic layout of graphs, diagrams, and networks.
\end{itemize}

JIProlog is a Java version of Prolog implementation and it can load and consume valid prolog files. JIProlog server supports socket communication. From the java end, a socket will be open on a particular port and wait for the client to connect. JIProlog server will be launched by loading a prolog file which will contain the core exshell prolog code and knowledge representation of data. JIProlog server is configured to communicate via socket and send/receive user query/output to and from Java application. When a JIProlog server finds a solution, an event is triggered and the solution is parsed from the Java end to visualize different results that have been provided by exshell. Yfiles has been used to visualize the tree diagram.

\section{Case Studies}
\begin{figure*}[ht]
\includegraphics[width=\textwidth]{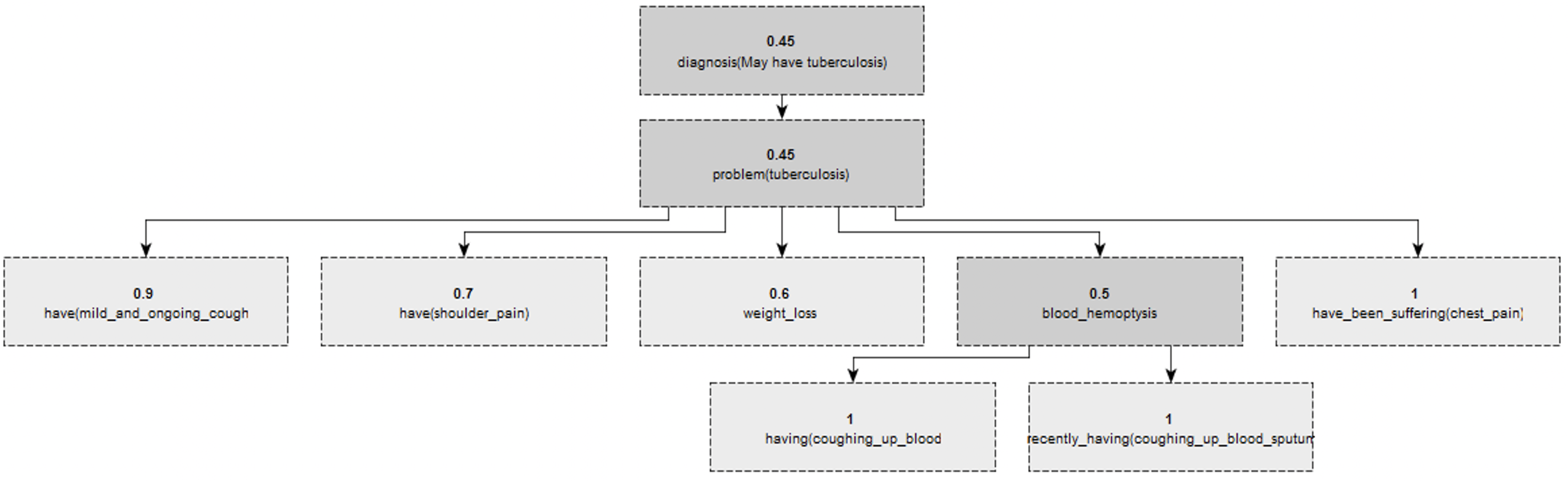}
\caption{Graphical visualization of diagnosis with certainty factor.} \label{fig_visual}
\end{figure*}

Fig.~\ref{fig_visual} shows an example of the diagnosis of Tuberculosis using our medical diagnosis expert system. Our system asks several question about symptoms of diseases. The inference engine reasons about the knowledge base and generates  decision making tree. The diagnosis result with certainty factor is visualized on the graphical user interface.

\begin{figure*}[ht]
\includegraphics[scale=0.7]{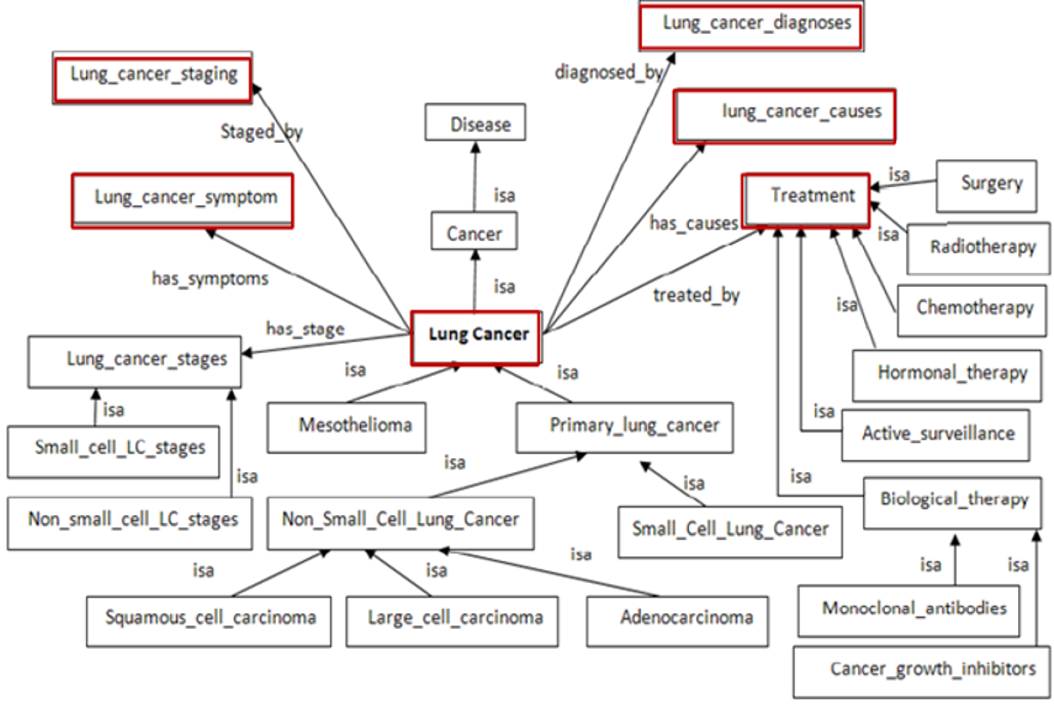}
\caption{Searching particular properties of lung cancer using our expert system.} \label{fig_case}
\end{figure*}

Fig.~\ref{fig_case} demonstrates an example of searching particular properties of lung cancer using our expert system. Our system has a simple semantic representation of diseases properties. In this system, several diseases functionalities such as disease details, treatments, hierarchical diseases information can be store in a semantic net. For example, lung cancer is a specific type of cancers, while mesothelioma and primary lung cancer are specific types of lung cancers, respectively. The treatment of the lung cancer can be surgery, radio therapy, chemotherapy, hormonal therapy, etc. By using knowledge representation for properties of diseases, our expert system can serve user's query to return a particular property of diseases and generate the knowledge tree visualization on the graphical user interface.  

\section{Conclusion}
In this paper, we design and implement a medical knowledge based expert system for disease diagnosis from symptoms. The system provides features for searching properties like symptoms, treatments, hierarchical clusters of particular diseases. The proposed disease diagnosis system uses a graphical user interface to facilitate users to interact with an expert system more easily. This system is generic, knowledge based, and can be integrated with any rule based expert system for diseases diagnosis. For future works, we plan to increase the capacity of our expert system by implementing more inference rules for semantic based knowledge representation and conduct real user study with patients and medical professionals. 

\section*{Acknowledgement}
This work was supported by a NVIDIA GPU Grant.

\nocite{*}
\bibliographystyle{ACM-Reference-Format}
\bibliography{sample-base}

\end{document}